\documentclass[11pt]{article}
\usepackage{coling2016}
\usepackage{times}
\usepackage{mdframed}
\usepackage{url}
\usepackage{latexsym}
\usepackage[normalem]{ulem} 
\usepackage{subfig}
\usepackage{wrapfig}
\usepackage[usenames,dvipsnames,table]{xcolor}
\usepackage{multirow}
\usepackage{colortbl}
\usepackage{amssymb}
\usepackage{paralist}
\usepackage{longtable}
\usepackage{enumitem}
\usepackage{multirow}
\usepackage{float}
\usepackage[pdftex]{graphicx}   
\usepackage{tcolorbox}
\usepackage{algorithm}
\usepackage{eqparbox,array}
\usepackage[noend]{algpseudocode}

\definecolor{airforceblue}{rgb}{0.36, 0.54, 0.66}
\definecolor{ao(english)}{rgb}{0.0, 0.5, 0.0}
\definecolor{dollarbill}{rgb}{0.52, 0.73, 0.4}
\definecolor{flame}{rgb}{0.89, 0.35, 0.13}

\usepackage{tikz}
\usetikzlibrary{arrows,calc,positioning,fit,backgrounds}
\definecolor{celadon}{rgb}{0.67, 0.88, 0.69}

\tikzstyle{intg}=[draw,minimum size=3em,text centered,text width=10.cm]
\tikzstyle{int}=[draw,minimum size=3em,text centered,text width=5.cm]
\tikzstyle{int1}=[draw,minimum size=2em,text centered,text width=3.cm]
\tikzstyle{int2}=[draw,minimum size=3em,text centered,text width=8.cm]
\usepackage{blindtext}
\usepackage[absolute]{textpos}
\usepackage{tabularx}
\newcolumntype{L}[1]{>{\raggedright\arraybackslash}p{#1}} 
\newcolumntype{C}[1]{>{\centering\arraybackslash}p{#1}} 
\newcolumntype{R}[1]{>{\raggedleft\arraybackslash}p{#1}} 
\newcolumntype{V}[1]{>{\centering\arraybackslash}m{#1}}

\begin{document}

\title{A Sentence Simplification System for Improving Relation Extraction}

\author{Christina Niklaus, Bernhard Bermeitinger, Siegfried Handschuh,  Andr\'{e} Freitas \\
  Faculty of Computer Science and Mathematics\\
  University of Passau\\
  Innstr. 41, 94032 Passau, Germany\\
  \scriptsize{{\tt \{christina.niklaus, bernhard.bermeitinger, siegfried.handschuh, andre.freitas \}}
  {\tt @uni-passau.de}}
  }

\date{}

\maketitle

\begin{abstract}
In this demo paper, we present a text simplification approach that is directed at improving the performance of state-of-the-art Open Relation Extraction (RE) systems. As syntactically complex sentences often pose a challenge for current Open RE approaches, we have developed a simplification framework that performs a pre-processing step by taking a single sentence as input and using a set of syntactic-based transformation rules to create a textual input that is easier to process for subsequently applied Open RE systems.
\end{abstract}

\section{Introduction}\blfootnote{This work is licensed under a Creative Commons Attribution 4.0 International Licence. Licence details:
http://creativecommons.org/licenses/by/4.0/}
Relation Extraction (RE) is the task of recognizing the assertion of relationships between two or more entities in NL text. Traditional RE systems have concentrated on identifying and extracting relations of interest by taking as input the target relations, along with hand-crafted extraction patterns or patterns learned from hand-labeled training examples. Consequently, shifting to a new domain requires to first specify the target relations and then to manually create new extraction rules or to annotate new training examples by hand \cite{Banko08}. As this manual labor scales linearly with the number of target relations, this supervised approach does not scale to large, heterogeneous corpora which are likely to contain a variety of unanticipated relations \cite{Schmid14}. To tackle this issue, \newcite{Banko08} introduced a new extraction paradigm named 'Open RE' that facilitates domain-independent discovery of relations extracted from text by not depending on any relation-specific human input.\\ 
Generally, state-of-the-art Open RE systems identify relationships between entities in a sentence by matching patterns over either its POS tags, e. g. \cite{Banko07,Fader11,Merhav12}, or its dependency tree, e. g. \cite{Nakas12,Mausam12,Xu13,Mesqu13}. However, particularly in long and syntactically complex sentences, relevant relations often span several clauses or are presented in a non-canonical form \cite{Angeli15}, thus posing a challenge for current Open RE approaches which are prone to make incorrect extractions - while missing others - when operating on sentences with an intricate structure.\\
To achieve a higher accuracy on Open RE tasks, we have developed a framework for simplifying the linguistic structure of NL sentences. It identifies components of a sentence which usually provide supplementary information that may be easily extracted without losing essential information. By applying a set of hand-crafted grammar rules that have been defined in the course of a rule engineering process based on linguistic features, these constituents are then disembedded and transformed into self-contained simpler context sentences. In this way, sentences that present a complex syntax are converted into a set of more concise sentences that are easier to process for subsequently applied Open RE systems, while still expressing the original meaning.

\section{System Description}
Referring to previous attempts at syntax-based sentence compression \cite{Dunl03,Zaji07,Pere13}, the idea of our text simplification framework is to syntactically simplify a complex input sentence by splitting conjoined clauses into separate sentences and by eliminating specific syntactic sub-structures, namely those containing only minor information. However, unlike recent approaches in the field of extractive sentence compression, we do not delete these constituents, which would result in a loss of background information, but rather aim at preserving the full informational content of the original sentence. Thus, on the basis of syntax-driven heuristics, components which typically provide mere secondary information are identified and transformed into simpler stand-alone context sentences with the help of paraphrasing operations adopted from the text simplification area.

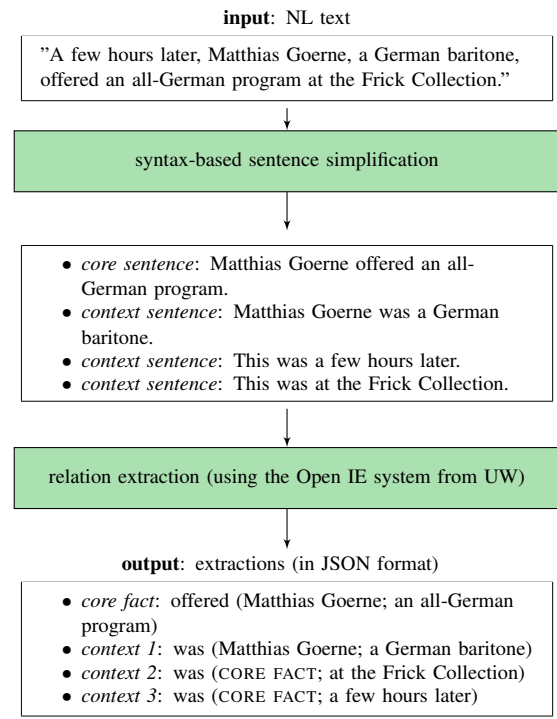
\begin{wrapfigure}{r}{7.5cm}
  \centering
    \begin{tikzpicture}[
      >=latex',
      scale=0.7,
      transform shape,
      auto
    ]
      
      \node [intg] (ki3) [draw = white] {\textbf{input}: NL text\\
      \begin{mdframed}
      "A few hours later, Matthias Goerne, a German baritone, offered an all-German program at the Frick Collection."
      \end{mdframed}};
      \node [intg] (ki4) [node distance=2cm,below of=ki3, fill=celadon] {syntax-based sentence simplification};
      \node [intg] (ki5) [draw = white, node distance=3cm,below of=ki4] {
      \begin{mdframed}
      \begin{compactitem}
      \setlength{\itemsep}{-1pt}
\item \textit{core sentence}: Matthias Goerne offered an all-German program.
\item \textit{context sentence}: Matthias Goerne was a German baritone.
\item \textit{context sentence}: This was a few hours later.
\item \textit{context sentence}: This was at the Frick Collection.

\end{compactitem}\end{mdframed}};
      
      \node [intg] (ki6) [node distance=3cm,below of=ki5, fill=celadon] {relation extraction (using the Open IE system from UW)};
      \node [intg] (ki8) [draw = white, node distance=3cm,below of=ki6] {\textbf{output}: extractions (in JSON format)
      \begin{mdframed}
      \begin{compactitem}
      \setlength{\itemsep}{-1pt}
\item \textit{core fact}: offered (Matthias Goerne; an all-German program)
\item \textit{context 1}: was (Matthias Goerne; a German baritone)
\item \textit{context 2}: was (\textsc{core fact}; at the Frick Collection)
\item \textit{context 3}: was (\textsc{core fact}; a few hours later)
\end{compactitem}
\end{mdframed}};

      \draw[->] (ki3) -- (ki4);
       \draw[->] (ki4) -- (ki5);
       \draw[->] (ki5) -- (ki6);
       \draw[->] (ki6) -- (ki8);

    \end{tikzpicture}
    \caption{Simplification and extraction pipeline}
    \label{pipeline}
\end{wrapfigure}

\paragraph{Definition of the Simplification Rules}
By analyzing the structure of hundreds of sample sentences from the English Wikipedia, we have determined constituents that commonly supply no more than contextual background information. These components comprise the following syntactic elements:
\vspace{0.1cm}
\begin{itemize}
\item \textbf{non-restrictive relative clauses} (e. g. \textit{"The city's top tourist attraction was the Notre Dame Cathedral, \uline{which welcomed 14 million visitors in 2013}."})
\item \textbf{non-restrictive} (e. g. \textit{"He plays basketball, \uline{a sport he participated in as a member of his high school's varsity team}."}) \textbf{and restrictive appositive phrases} (e. g. \textit{"He met with \uline{former British Prime Minister} Tony Blair."})
\item \textbf{participial phrases offset by commas} (e. g. \textit{"The deal, \uline{titled Joint Comprehensive Plan of Action}, saw the removal of sanctions."})
\item \textbf{adjective and adverb phrases delimited by punctuation} (e. g. \textit{"\uline{Overall}, the economy expanded at a rate of 2.9 percent in 2010."}) 
\item \textbf{particular prepositional phrases} (e. g. \textit{"\uline{In 2012}, Time magazine named Obama as its Person of the Year."})
\item \textbf{lead noun phrases} (e. g. \textit{"\uline{Six weeks later}, Alan Keyes accepted the Republican nomination."})
\item \textbf{intra-sentential attributions} (e. g. \textit{"\uline{He said that} both movements seek to bring justice and equal rights to historically persecuted peoples."})
\item \textbf{parentheticals} (e. g. \textit{"He signed the reauthorization of the State Children's Health Insurance Program \uline{(SCHIP)}."})
\end{itemize}
\vspace{0.1cm}
Besides, both conjoined clauses presenting specific features and sentences incorporating particular punctuation are disconnected into separate ones.\\
After having thus identified syntactic phenomena that generally require simplification, we have determined the characteristics of those constituents, using a number of syntactic features (constituency-based parse trees as well as POS tags) that have occasionally been enhanced with the semantic feature of NE tag. For computing them, a number of software tools provided by the Stanford CoreNLP framework have been employed (Stanford Parser, Stanford POS Tagger and Stanford Named Entity Recognizer).\footnote{\url{http://nlp.stanford.edu/software/}} Based upon these properties, we have then specified a set of hand-crafted grammar rules for carrying out the syntactic simplification operations which are applied one after another on the given input sentence. In that way, linguistically peripheral material is disembedded, thus producing a more concise core sentence which is augmented by a number of related self-contained contextual sentences (see the example displayed in figure \ref{pipeline}).

\begin{algorithm}
\caption{Syntax-based sentence simplification}\label{pseudocode}
\scriptsize
\begin{algorithmic}[1]
\Require sentence $s$
\Repeat
\State $r\gets$ next rule $\in$ R \Comment{Null if no more rules}
\If {$r$ is applicable to $s$ }
   \State $C$, $P$ $\gets$ apply $r_{extract}$ to $s$ \Comment{Identify the set of constituents $C$ to extract from $s$, and their positions $P$ in $s$}
\ForAll{constituents $c$ $\in$ $C$}
\State $context$ $\gets$ apply $r_{paraphrase}$ to $c$ \Comment{Produce a context sentence}
\State $contextSet$ $\gets$ add $context$ \Comment{Add it to the core's set of associated context sentences}
\EndFor
\EndIf
\Until{$R$ = $\emptyset$}
\State $core\gets$ delete tokens in $s$ at positions $p$ $\in$ $P$ \Comment{Reduce the input to its core}
\State \textbf{return} $core$ and $contextSet$\Comment{Output the core and its context sentences}
\end{algorithmic}
\end{algorithm}

\paragraph{Application of the Simplification Operations} The simplification rules we have specified are applied one after another to the source sentence, following a three-stage approach (see algorithm \ref{pseudocode}). First, clauses or phrases that are to be separated out - including their respective antecedent, where required - have to be identified by pattern matching. In case of success, a context sentence is constructed by 
either linking the extractable component to its antecedent or by inserting a complementary constituent that is required in order to make it a full sentence. Finally, the main sentence has to be reduced by dropping the clause or phrase, respectively, that has been transformed into a stand-alone context sentence.\\
In this way, a complex source sentence is transformed into a simplified two-layered representation in the form of core facts and accompanying contexts, thus providing a kind of normalization of the input text. Accordingly, when carrying out the task of extracting semantic relations between entities on the reduced core sentences, the complexity of determining intricate predicate-argument structures with variable arity and nested structures from syntactically complex input sentences is removed. Beyond that, the phrases of the original sentence that convey no more than peripheral information are converted into independent sentences which, too, can be more easily extracted under a binary or ternary predicate-argument structure (see the example illustrated in figure \ref{pipeline}).

\section{Evaluation}

\begin{wrapfigure}{r}{11cm}
  \centering
    \begin{tikzpicture}[
      >=latex',
      scale=0.75,
      transform shape,
      auto
    ]
      
      \node [int1] (ki3) [draw = black, fill = gray] {Matthias Goerne};
      \node [int] (ki4) [node distance=8cm,right of=ki3, draw=white] {an all-German program};
      \node [int] (ki5) [node distance=2cm,below of=ki3, draw= white] {a German baritone};
      \node (set d) [left=of ki3,xshift=-0.85cm] {}; 
      \node (set c) [right=of ki4,xshift=0.3cm] {}; 
      \node[draw, fit=(set d)(ki3)(ki4)(set c), fill=lightgray] (ki) {};
      \node [int1] (ki3) [draw = black, fill = gray] {Matthias Goerne};
      \node [int] (ki4) [node distance=8cm,right of=ki3, draw=none] {an all-German program};
      \node [int] (ki6) [node distance=2cm,below of=ki, draw= white] {at the Frick Collection};
       \node [int] (ki7) [node distance=2cm,below of=ki4, draw= white] {a few hours later};

       \draw[->] (ki3) -- (ki4) node [midway] {offered};
       \draw[->] (ki3) -- (ki5) node [midway] {was};
      
      \draw[->] (ki) -- (ki6) node [midway] {was};
       \draw[->] (ki) -- (ki7) node [midway] {was};
       
    \end{tikzpicture}
    \caption{Extracted relations when operating on the simplified sentences}
    \label{example1}
    \end{wrapfigure}
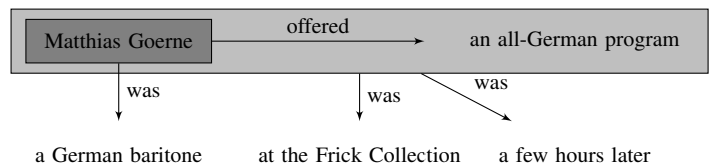
   
\begin{figure}[!ht]
    \centering
    \begin{tikzpicture}[
      >=latex',
      scale=0.85,
      transform shape,
      auto
    ]
      
      \node [int] (ki3) [draw = white] {\fbox{Matthias Goerne}};
      \node [int2] (ki4) [node distance=2cm,below of=ki3, draw=white] {\fbox{an all-German program at the Frick Collection}};

       \draw[->] (ki3) -- (ki4) node [midway] {offered};

    \end{tikzpicture}
    \caption{Result without a prior simplification step}
    \label{example2}
\end{figure}
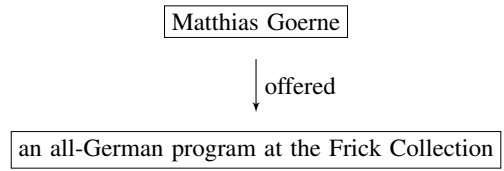

   The results of an experimental evaluation show that state-of-the-art Open RE approaches obtain a higher accuracy and lower information loss when operating on sentences that have been pre-processed by our simplification framework. In particular when dealing with sentences that contain nested structures, Open RE systems benefit from a prior simplification step (see figures \ref{example1} and \ref{example2} for an example). The full evaluation methodology and detailed results are reported in \newcite{Nikl16}.

\section{Usage}
The text simplification framework is publicly available\footnote{\url{https://gitlab.com/nlp-passau/SimpleGraphene}} as both a library and a command line tool whose workflow is depicted in figure \ref{pipeline}. It takes as input NL text in the form of either a single sentence or a file with line separated sentences. As described above, each input sentence is first transformed into a structurally simplified version consisting of 1 to $n$ core sentences and 0 to $m$ associated context sentences. In a second step, the relations contained in the input are extracted by applying the Open IE system\footnote{\url{https://github.com/allenai/openie-standalone}} upon the simplified sentences. Finally, the results generated in this way are written to the console or a specified output file in JSON format.
As an example, the outcome produced by our simplification system when applied to a full Wikipedia article is provided online.\footnote{\url{https://gitlab.com/nlp-passau/SimpleGraphene/tree/master/examples}}

\section{Conclusion}
We have described a syntax-driven rule-based text simplification framework that simplifies the linguistic structure of input sentences with the objective of improving the coverage of state-of-the-art Open RE systems. As an experimental analysis has shown, the text simplification pre-processing improves the result of current Open RE approaches, leading to a \textit{lower information loss} and a \textit{higher accuracy} of the extracted relations.

\bibliographystyle{acl}
\bibliography{coling2016}

\end{document}